\documentclass[runningheads]{llncs}

\usepackage{amsmath,amsfonts,amssymb,amsthm,mathrsfs,bm}
\usepackage{graphicx}
\usepackage{subfigure}
\usepackage{booktabs}
\usepackage{multirow}
\usepackage{color}

\def\etal{\emph{et al.}}

\begin{document}
\title{Robust Multi-view Registration of Point Sets with Laplacian Mixture Model}
%
%
\author{Jin Zhang\inst{1}
    \and
    Mingyang Zhao\inst{2,3}
    \and
    Xin Jiang\inst{1,4} \thanks{Corresponding author}
    \and
    Dong-Ming Yan\inst{2}}
\authorrunning{J. Zhang et al.}
%
\institute{Key Laboratory of Mathematics, Informatics and Behavioral Semantics, School of Mathematical Science, Beihang University, Beijing, China \\ \email{jiangxin@buaa.edu.cn}\and
    National Key Laboratory of Pattern Recognition, Institute of Automation, Chinese Academy of Sciences (CAS), Beijing, China
    \and Beijing Academy of Artificial Intelligence, Beijing, China \and Peng Cheng Laboratory, Shenzhen, Guangdong, China\\
}
\maketitle              
\begin{abstract}
    Point set registration is an essential step in many computer vision applications, such as 3D reconstruction and SLAM. Although there exist many registration algorithms for different purposes, however, this topic is still challenging due to the increasing complexity of various real-world scenarios, such as heavy noise and outlier contamination. In this paper, we propose a novel probabilistic generative method to simultaneously align multiple point sets based on the heavy-tailed  Laplacian distribution. The proposed method assumes each data point is generated by a \emph{Laplacian Mixture Model} (LMM), where its centers are determined by the corresponding points in other point sets. Different from the previous \emph{Gaussian Mixture Model} (GMM) based method, which minimizes the quadratic distance between points and centers of Gaussian probability density, LMM minimizes the sparsity-induced $L_1$ distance, thereby it is more robust against noise and outliers.
    We adopt \emph{Expectation-Maximization} (EM) framework to solve LMM parameters and rigid transformations. We approximate the $L_1$ optimization as a linear programming problem by exponential mapping in Lie algebra, which can be effectively solved through the \emph{interior point method}. To improve efficiency, we also solve the $L_1$ optimization by \emph{Alternating Direction Multiplier Method} (ADMM). We demonstrate the advantages of our method by comparing it with representative state-of-the-art approaches on benchmark challenging data sets, in terms of robustness and accuracy.

    \keywords{Point set registration \and $L_1$ optimization \and GMM \and LMM.}
\end{abstract}
\section{Introduction}
Point set registration is a fundamental problem that has wide applications in computer vision~\cite{yang2015go}, robotics~\cite{yang2020teaser}, computer graphics~\cite{huang2020uncertainty}, medical image analysis~\cite{rasoulian2012group} and so on. With the advent of sensors such as LiDAR (Light Detection and Ranging) and depth cameras, it becomes relatively easy to capture real-world 3D scene data. However, one usually attains partial point clouds at once due to the 3D nature of the objects. To accurately reconstruct the 3D model, it is necessary to align multiple point clouds acquired from multi views of the object into a unified coordinate system.

Point set registration has been extensively studied in literature, and many efforts are devoted to the pair-wise (two sets) registration problem. Among these methods, the most classical one is the \emph{Iterative Closest Point} (ICP) algorithm~\cite{paulj1992method}, in which the registration is decomposed as the alternative implementation of point correspondence and transformation estimation. Given a set of data points, the rigid transformation of ICP is solved by minimizing the summation of the squared distance of the closest point pairs. However, ICP tends to get trapped in a local optimum because of the hard assignment scheme, improper initialization, occlusions, and the interference of noise and outliers.

In contrast to the hard assignment scheme of ICP, previous works also explore the use of statistical models for registration, such as GMM~\cite{jian2010robust}, which replaces the previous binary assignment with probability. These methods formulate point set registration as a probability density estimation, where the mean of each component is initialized as the point location. In principle, probability registration can provide better estimation for convergence and geometric matching, thereby they are further extended for multi-view registration~\cite{evangelidis2017joint}. Most existing probability methods rely on GMM for registration, nevertheless, Gaussian distribution minimizes the quadratic distance between the data points and their means, which makes GMM susceptible to noise with heavy tail and sensitive to outliers~\cite{gao2008robust}.

In this paper, to address the aforementioned problems, we propose a novel and robust multi-view registration method based on the LMM composed by the Laplacian distribution. Due to the heavy-tail property and the sparsity-induced $L_1$ norm, LMM is more robust against outliers. We formulate point set registration as a likelihood estimation problem, which can be solved by EM framework. To handle the $L_1$ optimization, we customize the rotation estimation to a linear programming problem by exponential mapping and name it as \emph{linear programming approximation} (LPA). Moreover, inspired by~\cite{bouaziz2013sparse}, we also use ADMM to solve $L_1$ optimization, which has higher efficiency without significant accuracy decreasing. We test and compare the proposed methods (including LPA and ADMM) with representative state-of-the-art approaches in terms of accuracy and robustness, and the results indicate that our methods outperform previous ones with higher accuracy and are more robust against noise and outliers.

In a nutshell, the main contributions of this paper are twofold as follows:
\begin{itemize}
    \item We propose a novel multi-view registration method for point clouds based on the sparsity-induced Laplacian mixture model. Due to the $L_1$ norm of Laplacian distribution, the proposed method is more robust against noise and outliers;
    \item To handle $L_1$ optimization, we customize it as a linear programming problem via exponential mapping, which can be effectively solved by the interior point method. For efficiency, we further deduce it into the ADMM framework, thereby it can also be solved by the ADMM algorithm.
\end{itemize}

\section{Related Work}
We briefly review the related work of point set registration from the perspective of pair-wise registration and multi-view registration.
\subsection{Pair-wise Registration}
The objective of pair-wise registration is to align two point sets by solving a transformation matrix. The most popular algorithm of pair-wise registration is ICP~\cite{paulj1992method}, which composes two iterative steps: 1) correspondence step in which the point correspondence between two point sets is established, and 2) transformation step in which the transformation matrix based on the current correspondence is updated. ICP takes the least-squares estimator as the objective function, thereby the closed-form solution in each step can be attained through \emph{singular value decomposition} (SVD). However, as the Gauss-Markov theorem~\cite{rousseeuw2005robust} pointed, the least-squares estimator is sensitive to outliers. Moreover, ICP requires good initialization and fails to handle non-overlapping point sets. To improve performance, many variants of ICP have been proposed. Fitzgibbon~\cite{fitzgibbon2003robust} adopts M-estimator for registration error minimization and solves the problem by non-linear Levenberg-Marquardt optimization, thereby proper initialization is required. Chetverikov~\etal~\cite{chetverikov2002trimmed} propose a trim scheme named TrICP to automatically remove non-overlapping regions for accurate registration. For efficiency, Rusu \etal~\cite{rusu2009fast} introduce Fast Point Feature Histograms (FPFH) to describe the local geometry around a point to reduce the computational complexity. Lei~\etal~\cite{lei2017fast} compute eigenvalues and normals from multiple scales and take them as local descriptors to speed up matching. However, descriptor-based methods~\cite{lei2017fast,guo2020learning} are susceptible to point sets with noise and low overlapping. Recently, deep learning-based methods such as~\cite{aoki2019pointnetlk} are also proposed to estimate the transformation matrix. Nevertheless, the lack of comprehensive registration datasets results in a hard time for the learning methods to grasp all shape variations.

Alternatively, probability registration methods adopt GMM  to represent data points, such as Robust Point Matching~\cite{chui2003new}, Coherent Point Drift~\cite{myronenko2010point}, and GMMReg~\cite{jian2010robust}. These methods either simultaneously model the source and the target point sets by GMM, and the transformation matrix is solved by minimizing the discrepancy between the two GMMs, or singly model the source point set as GMM, and then evoke maximum likelihood estimation to fit the target point set. However, GMM fails to attain accurate registration results under the contamination of noise with a heavy tail. Moreover, since GMM minimizes the quadratic distance between the points and its means, it suffers from severe outliers. In contrast, we introduce Laplacian mixture models to model data distribution, which adopt the $L_1$ norm for error evaluation, thereby it is more robust against outliers for registration.
\subsection{Multi-view registration}
Multi-view registration aims to simultaneously register multiple point sets from different views, which is often solved through sequential pair-wise registration. Transformation parameters are sequentially updated by ICP or probability methods if a new point set is added. Except for the drawbacks from pair-wise registration, these methods also suffer from error accumulation and propagation. Bergevin~\etal~\cite{bergevin1996towards} propose a star-network and sequentially put one point set in the center of it, then pair-wise registration by ICP is implemented to align the central point set with the other ones. In contrast,  Williams~\etal~\cite{williams2001simultaneous} simultaneously compute the correspondence between all point sets, which is time-consuming. Mateo~\etal~\cite{mateo2014bayesian} introduce the Bayesian perspective to assign different weights for different correspondences, with the detection of false correspondences. Although the registration accuracy is improved, it needs to compute many variables.

Additionally, information theoretic measures are also customized for multi-view registration task. Wang~\etal~\cite{wang2006groupwise} first represent point sets by cumulative distribution function (CDF), and then minimize the Jensen-Shannon divergence between CDFs for multi-view registration. Later, Chen~\etal~\cite{chen2010group} use the Havrda-Charvat divergence to evaluate the differences between CDFs, compared with previous Jensen-Shannon divergence, this method is more efficient. Nevertheless, information theory based approaches are still with low efficiency.

Recently, GMM is generalized for multi-view registration by~\cite{evangelidis2017joint} named JRMPC, which assumes  data points are generated by a central GMM, and then casts the registration task as a clustering process.
By this, global information of point sets is combined to avoid the error accumulation.
Zhu \etal ~\cite{zhu_registration_2020} propose a method named EMPMR, which assumes that each data point is generated from a GMM whose Gaussian centroids are composed of corresponding points from other point sets.
Nevertheless, GMM suffers from heavy-tail noise, and is sensitive to severe outliers due to the $L_2$ norm.
Based on JMRPC, TMM~\cite{ravikumar_group-wise_2018}  achieves better robustness by replacing the Gaussian distribution with a $t$-distribution, however, its time consumption is relatively considerable.
To address these drawbacks, we propose a novel and robust multi-view registration method based on the sparsity-induced $L_1$ norm, which is effectively solved by LPA or ADMM.


\section{Methodology}

\subsection{Multivariate Laplacian Distribution}

Suppose $ \mathbf{x}\in \mathbb{R}^{d}$  is a random variable following the multivariate Laplacian distribution, with the probability density function as
\begin{equation}
    \mathcal{L}\left( \boldsymbol{x};\boldsymbol{\mu },b \right) =\frac{1}{\left( 2b \right) ^d}\exp \left( -\frac{\left\| \boldsymbol{x}-\boldsymbol{\mu } \right\| _1}{b} \right),
\end{equation}
where $ \boldsymbol{\mu }$ and $b$ represent the mean and the scale parameter of the $d$-dimensional Laplacian distribution, respectively, and $\left\|\cdot\right\|_1$ denotes the sparsity-induced $L_1$ norm, summarizing the absolute values of all elements of a vector.
Previous work has shown that the Gaussian distribution is sensitive to heavy outliers because of its short tails~\cite{gao2008robust,azam2020multivariate}, while the Laplacian distribution has heavier tails.

\subsection{Laplacian Mixture Model}
Let $\mathcal{X} =\left\{ \boldsymbol{X}_i \right\} _{i=1}^{M}$ be the union of  $ M $ point sets, and $\boldsymbol{X}_{i}=\left[\boldsymbol{x}_{i1},\boldsymbol{x}_{i2},\ldots,\boldsymbol{x}_{iN_{i}}\right] \in \mathbb{R}^{3 \times N_{i}}$ be the $ i $-th point set, where $ \boldsymbol{x}_{il} $ is the $ l $-th point of $ \boldsymbol{X}_{i} $ and $ N_{i} $ is the cardinality of $ \boldsymbol{X}_{i} $. The task of multi-view registration is to align the multiple point sets in $\mathcal{X}$ to the same center frame. To this end, we solve the rotation matrix $ \boldsymbol{R}_i $ and the translation vector $ \boldsymbol{t}_i$ for each $ \boldsymbol{X}_{i}$, and represent the transformed coordinates as $\boldsymbol{\hat{x}}_{il}=\boldsymbol{R}_i\boldsymbol{x}_{il}+\boldsymbol{t}_i $. Due to the influence of noise, it is hard to hope that corresponding points in different frames will have the same coordinate after alignment. In contrast, we assume the multi-view point sets will constitute multiple clusters.

Suppose each data point $ \boldsymbol{\hat{x}}_{il} $ is generated from a unique LMM, where the Laplacian centers are composed of the corresponding points in other frames after alignment. However, it is difficult to directly get the accurate correspondence relationship between different point sets. Here we adopt the \emph{nearest neighbor search} based on the \emph{kd-tree} to approximate the corresponding points. In specific, for point $ \boldsymbol{\hat{x}}$ of the center frame, we denote $ {c}_{j}(\boldsymbol{\hat{x}}) $ as the nearest neighbor of $\boldsymbol{\hat{x}}$ from the $ j $-th point set, which can be defined as:
\begin{equation}
    c_j\left( \boldsymbol{\hat{x}} \right) =\underset{\{\boldsymbol{\hat{x}}_{jh}\}_{h=1}^{N_j}}{\mathrm{arg}\min}\left\| \boldsymbol{\hat{x}}-\boldsymbol{\hat{x}}_{jh} \right\| _1.
\end{equation}
For simplicity, we use isotropic covariance $b$ and equal membership probability for all LMM components. Thereby, the LMM of $ \boldsymbol{x}\in \boldsymbol{X}_i $  is defined as
\begin{equation}\label{prob_eq}
    P\left( \boldsymbol{x} \right) =\sum_{j\ne i}^M{\frac{1}{M-1}}\mathcal{L} \left( \boldsymbol{R}_i\boldsymbol{x}+\boldsymbol{t}_i;{c}_{j}(\boldsymbol{\hat{x}}),b \right).
\end{equation}Note that different from previous GMM based methods, in which a uniform distribution has to be added to account for noise and outliers, there is no need for our LMM based method, since Laplacian distribution has a heavy tail and is sparse enough to accommodate outliers.

We adopt the \emph{maximum likelihood estimation} (MLE) to solve unknown parameters. The log-likelihood function of observed data points is
\begin{equation}
    L(\Theta;\mathcal{X} )=\log P(\mathcal{X}\mid \Theta)
    =\sum_{i=1}^{M}\sum_{l=1}^{N_i}
    \log\left( \sum_{j\ne i}^M{\frac{1}{M-1}}\mathcal{L} \left( \boldsymbol{\hat{x}}_{il};{c}_{j}(\boldsymbol{\hat{x}}_{il}),b \right) \right),
\end{equation}where $\Theta =\left\{ \left\{ \boldsymbol{R}_i,\boldsymbol{t}_i \right\} _{i=1}^{M},b \right\}$ represents all parameters. It is intractable to directly solve the maximum likelihood solution for such a complex model. Instead, we adopt the effective  EM framework by introducing latent variables for solving, as presented in the following.

\section{Registration by EM Algorithm}
To estimate the parameters by EM algorithm,
we first define a set of latent variables $\mathcal{Z} =\left\{ Z_{il}\mid 1\le i\le M,~1\le l\le N_i \right\}$, where $ Z_{il} = j $ means that  $ \boldsymbol{\hat{x}}_{il} $  is generated from the $ j $-th component of the LMM.
Given all observed data $\mathcal{X} $, model parameters can be estimated by maximizing the expectation of the log-likelihood function:
\begin{equation}\label{expect_eq}
    \begin{aligned}
        \mathcal{E} \left( \Theta ;\mathcal{X} ,\mathcal{Z} \right) & =\mathbb{E} _{\mathcal{Z}}\left[ \log P\left( \mathcal{X} ,\mathcal{Z} ;\Theta \right) \right] =\sum_Z{P}\left( Z\mid \mathcal{X} ;\Theta \right) \log P\left( \mathcal{X} ,Z;\Theta \right) \\
                                                                    & =\sum_Z{P}\left( Z\mid \mathcal{X} ;\Theta \right) \left( \log P\left( \mathcal{X} \mid Z;\Theta \right) +\log P\left( Z;\Theta \right) \right).                                             \\
    \end{aligned}
\end{equation}
Since we regard each Laplacian component equal, the prior probability $ P(Z;\Theta) $  is a constant term. In addition, $P(\mathcal{X} \mid Z;\Theta)$ can be derived from Eq.~\eqref{prob_eq}, namely,
$P\left(\boldsymbol{x}_{il}\mid Z_{il}=j;\Theta  \right)=\mathcal{L}\left( \boldsymbol{\hat{x}}_{il};c_j\left( \boldsymbol{\hat{x}}_{il} \right),b\right).$

After ignoring constant terms, the objective function of Eq.~(\ref{expect_eq}) is reformulated as

\begin{equation}
    \begin{aligned}
        f\left( \Theta \right)
         & =\sum_{i=1}^{M}\sum_{l=1}^{N_i} 	\sum_{j\ne i}^M 	\alpha_{ilj}\log\mathcal{L} \left( \boldsymbol{\hat{x}}_{il};{c}_{j}(\boldsymbol{\hat{x}}_{il}),b \right) \\
         & =-\sum_{i,l,j}\alpha_{ilj}\left(\frac{1}{b}
        \left\| \boldsymbol{\hat{x}}_{il}-{c}_{j}(\boldsymbol{\hat{x}}_{il}) \right\|_1 +d\log 2b \right),
    \end{aligned}
\end{equation}
where $ \alpha_{ilj}=P(Z_{il}=j\mid\boldsymbol{x}_{il};\Theta)$ denotes the posterior and $ d $ denotes the data dimension ($d=3$ in our case). Therefore, the multi-view registration problem is cast into a constrained optimization problem as follows:
\begin{equation}\label{opt1}
    \begin{array}{l}
        \hat{\Theta}=\underset{\Theta}{\mathrm{arg}\max} \,\,f(\Theta), ~
        \,\,\mathrm{s}.\mathrm{t}. \,\,
        \boldsymbol{R}_{i}\in \mathrm{SO}(3),\quad \forall i\in [1,...,M] \\
    \end{array}.
\end{equation}

In order to maximize $f(\Theta)$ by EM algorithm, we alternatively perform E-step and M-step as follows, after which, the transformation parameters of multi-view registration are attained.

\subsection{E-Step}
Given data point set $\mathcal{X}$ and currently
estimated parameters $ \Theta^{(k)} $, E-step calculates the distribution of latent variable $\mathcal{Z}$, which can be divided into two steps. The first step is to establish point correspondence.
For point $ \boldsymbol{x}_{il} $ in $ i $-th point set, it is initially transformed into the
center frame with new coordinate
$ \boldsymbol{\hat{x}}_{il}^{(k)}=\boldsymbol{R}_i^{(k)}\boldsymbol{x}_{il}+\boldsymbol{t}_i^{(k)} $.
Then the corresponding point in $j$-th frame can be approximated by the nearest neighbour of $ \boldsymbol{\hat{x}}_{il}^{(k)}$:
\begin{equation}
    c_j\left( \boldsymbol{\hat{x}}_{il}^{(k)} \right) =
    \mathop{\arg \min}_{\{\boldsymbol{\hat{x}}_{jh}\}_{h=1}^{N_j}}
    \left\| \boldsymbol{\hat{x}}_{il}^{(k)}-\boldsymbol{\hat{x}}_{jh}^{(k)} \right\| _1.
\end{equation}
The second step updates $\alpha_{ilj}^{(k+1)}=P(Z_{il}=j\mid\boldsymbol{x}_{il};\Theta^{(k)})$, which represents the posterior probability of point $\boldsymbol{x}_{il}$
generated from $j$-th component of LMM.
According to Bayesian formula, we have

\begin{equation}
    \begin{aligned}
        \alpha_{ilj}^{(k+1)}
         & =  \frac{\mathcal{L}\left(\boldsymbol{\hat{x}}_{il}^{(k)};c_j( \boldsymbol{\hat{x}}_{il}^{(k)} ),b^{(k)}\right)}
        {\sum_{j\ne i}^M{\mathcal{L}\left(\boldsymbol{\hat{x}}_{il}^{(k)};c_j( \boldsymbol{\hat{x}}_{il}^{(k)} ),b^{(k)}\right)}} =  \cfrac{\beta _{ilj}^{(k+1)}}{\sum_{j\ne i}^M{\beta _{ilj}^{(k+1)}}},
    \end{aligned}
\end{equation}
where $\beta _{ilj}$ denotes the probability density of Laplacian distribution:
\begin{equation}
    \beta _{ilj}^{(k+1)}=\frac{1}{\left( 2b^{(k)} \right) ^d}\exp \left( -\frac{\left\|
            \boldsymbol{\hat{x}}_{il}^{(k)}-c_j( \boldsymbol{\hat{x}}_{il}^{(k)} )
            \right\| _1}{b^{(k)}} \right).
\end{equation}

\subsection{M-Step}
M-step estimates the parameters $\Theta^{(k+1)}$ by maximizing the expectation $f(\Theta)$ with current values $\alpha_{ilj}^{(k+1)}$ and  $c_j(\boldsymbol{\hat{x}}_{il}^{(k)})$. Since it is difficult to directly estimate all parameters in $\Theta =\{ \{ \boldsymbol{R}_i,\boldsymbol{t}_i \} _{i=1}^{M},b \}$, we first solve transformation parameters  $\{\boldsymbol{R}_i,\boldsymbol{t}_i\}$ and then solve the variance scale $b$. Transformation parameters are estimated by
\begin{equation}\label{opt_i}
    \left\{ \begin{aligned}
        \mathop{\arg\min}_{\boldsymbol{R}_i,\boldsymbol{t}_i}
        \quad                        &
        \sum_{l=1}^{N_i}{\sum_{j\ne i}^M{\alpha _{ilj}^{(k+1)}}}\left\| \boldsymbol{R}_i\boldsymbol{x}_{il}+\boldsymbol{t}_i-c_j(\boldsymbol{\hat{x}}_{il}^{(k)}) \right\| _1 \\
        \mathrm{s}.\mathrm{t}. \quad & \boldsymbol{R}_{i}^{T}\boldsymbol{R}_i=\boldsymbol{I}\,\,\,\mathrm{and}\,\left| \boldsymbol{R}_i \right|=1.                            \\
    \end{aligned} \right.
\end{equation}
The above equation is a \emph{weighted least absolute value} (WLAV) problem with $\mathrm{SO}(3)$ constraint. Although there is lack of a closed-form solution, we propose two methods to iteratively solve it in Section~\ref{sec:WLAV}. After updating transformation parameters, we take the partial derivative of $f(\Theta)$ with respect to $b$ and equate it to zero, then the update of $b^{( k+1 )}$ is
\begin{equation}
    b^{\left( k+1 \right)}=\frac{\sum_{i,j,l}{\alpha _{ilj}^{\left( k+1 \right)}\left\| \boldsymbol{\hat{x}}_{il}^{\left( k+1 \right)}-c_j(\boldsymbol{\hat{x}}_{il}^{\left( k+1 \right)}) \right\| _1}}{d\sum_i{N_i}},
\end{equation}
where $N_i$ denotes the cardinality of the $i$-th point set. As can be seen, $b$ presents the weighted average of the deviations.

\subsection{WLAV subproblem}\label{sec:WLAV}
Previous GMM based approaches usually need to solve a weighted least square problem, and the closed-form solution can be attained through SVD. However, due to the $L_1$ norm optimization, WLSV problem requires iterative solving. In this work, we propose two methods to solve this kind of problem, namely, LPA based and ADMM based methods, as presented in the following.

\subsubsection{LPA Method}
We first transform the sub-problem \eqref{opt_i} into a canonical form:
\begin{equation}\label{std_opt}
    \min_{\boldsymbol{R},\boldsymbol{t}} \quad \sum_{i=1}^n{w_i\left\| \boldsymbol{Rp}_i+\boldsymbol{t}-\boldsymbol{q}_i \right\| _1}\quad
    \mathrm{s}.\mathrm{t}.~\boldsymbol{R}\in \mathrm{SO(}3),\\
\end{equation}
where $\boldsymbol{P}=[\boldsymbol{p}_1,\ldots,\boldsymbol{p}_n]\in \mathbb{R} ^{3\times n}$ and $\boldsymbol{Q}=[\boldsymbol{q}_1,\ldots,\boldsymbol{q}_n]\in \mathbb{R} ^{3\times n}$ represent the source point set and the target point set respectively, while  $w_i$ represents the weight. By exponential mapping, the rotation matrix $\boldsymbol{R}$ can be written as:
\begin{equation}
    \boldsymbol{R}=\exp \left( \left[ \boldsymbol{r} \right] _{\times} \right) =\boldsymbol{I}+\left[ \boldsymbol{r} \right] _{\times}+\frac{1}{2!}\left[ \boldsymbol{r} \right] _{\times}^{2}+\frac{1}{3!}\left[ \boldsymbol{r} \right] _{\times}^{3}+\cdots ,
    \label{eq:expand}
\end{equation}
where $\left[ \boldsymbol{r} \right] _{\times}$ represents the skew-symmetric
matrix of $\boldsymbol{r}\in \mathbb{R}^3$, expressed as
\begin{equation}
    \left[ \boldsymbol{r} \right] _{\times}=\left[ \begin{matrix}
            0    & -r_3 & r_2  \\
            r_3  & 0    & -r_1 \\
            -r_2 & r_1  & 0    \\
        \end{matrix} \right] .
\end{equation}
To eliminate the $\mathrm{SO}(3)$ constraint,
we linearly approximate the rotation matrix $\boldsymbol{R}$ by neglecting the higher-order terms of (\ref{eq:expand}). Then the deviation can be approximated as
\begin{equation}\label{linear_deviation}
    \boldsymbol{Rp}-\boldsymbol{q}+\boldsymbol{t}\approx \left[ \begin{matrix}
            -\left[ \boldsymbol{p} \right] _{\times} & \boldsymbol{I}_{3\times 3} \\
        \end{matrix} \right] \left[ \begin{array}{c}
            \boldsymbol{r} \\
            \boldsymbol{t} \\
        \end{array} \right] +\boldsymbol{p}-\boldsymbol{q},
\end{equation}since $\left[ \boldsymbol{r} \right] _{\times}\boldsymbol{p}=-\left[ \boldsymbol{p} \right] _{\times}\boldsymbol{r}.$

Then the objective function can be reformulated as:
\begin{footnotesize}
    \begin{equation}
        \begin{aligned}
            \sum_{i=1}^N{w_i\left\| \boldsymbol{Rp}_i-\boldsymbol{q}_i+\boldsymbol{t} \right\| _1}
            \approx \left\| \left[ \begin{matrix}
                    -w_1\left[ \boldsymbol{p}_1 \right] _{\times} & \boldsymbol{I}_3 \\
                    \vdots                                        & \vdots           \\
                    -w_N\left[ \boldsymbol{p}_N \right] _{\times} & \boldsymbol{I}_3 \\
                \end{matrix} \right] \left[ \begin{array}{c}
                    \boldsymbol{r} \\
                    \boldsymbol{t} \\
                \end{array} \right] -\left[ \begin{array}{c}
                    -w_1\left( \boldsymbol{p}-\boldsymbol{q} \right) \\
                    \vdots                                           \\
                    -w_N\left( \boldsymbol{p}-\boldsymbol{q} \right) \\
                \end{array} \right] \right\| _1
            =\left\| \boldsymbol{Ax}-\boldsymbol{b} \right\| _1,
        \end{aligned}
    \end{equation}
\end{footnotesize}where $\boldsymbol{x}$ denotes $\left[\boldsymbol{r}, \boldsymbol{t} \right]^{T}$. Let $\boldsymbol{u}=|\boldsymbol{Ax}-\boldsymbol{b}|$, the original WLAS problem
is transformed into a linear programming problem:
\begin{equation}
    \min_{\boldsymbol{u},\boldsymbol{x}} \quad \sum_{i=1}^N{u_i},\quad
    \mathrm{s}.\mathrm{t}.~ -\boldsymbol{u}+\boldsymbol{Ax}-\boldsymbol{b}\le \mathbf{0}~ \text{and}~
    -\boldsymbol{u}-\boldsymbol{Ax}+\boldsymbol{b}\le \mathbf{0},\\
\end{equation}
which can be efficiently solved by interior point methods~\cite{kim2007interior}. Moreover, we deduce another method to solve $L_1$ optimization based on the ADMM in the following. We name the two optimizations as Ours-LPA and Ours-ADMM, respectively, and compare them with previous approaches in Section~\ref{sec:experiment}.

\subsubsection{ADMM Method}\label{sec:ADMM}
Considering the constraint $\boldsymbol{z}_i=\boldsymbol{w}_i\left( \boldsymbol{R}\boldsymbol{p}_i+\boldsymbol{t}  -\boldsymbol{q}_i\right)$, Eq.~\eqref{std_opt} can be reformulated as:
\begin{equation}
    \begin{aligned}
        \min_{\boldsymbol{R},\boldsymbol{t}} & \quad \sum_{i=1}^n{\left\| \boldsymbol{z}_i \right\| _1}                                                                  \\
        \mathrm{s}.\mathrm{t}.               & \quad \boldsymbol{R}\in \mathrm{SO(}3),                                                                                   \\
                                             & \quad \boldsymbol{z}_i=\boldsymbol{w}_i\left( \boldsymbol{Rp}_i+\boldsymbol{t}-\boldsymbol{q}_i \right) ,\quad i=1,...,n. \\
    \end{aligned}
\end{equation}
Then the augmented Lagrangian function of ADMM is
\begin{equation}
    L_{\rho}\left( \boldsymbol{R},\boldsymbol{t},\boldsymbol{z},\lambda \right) =\sum_{i=1}^n{\left( \left\| \boldsymbol{z}_i \right\| _1+\frac{\rho}{2}\left\| \boldsymbol{z}_i-\boldsymbol{s}_i+\frac{1}{\rho}\boldsymbol{\lambda }_i \right\| _{2}^{2}-\frac{1}{2\rho}\left\| \boldsymbol{\lambda }_i \right\| _{2}^{2} \right)},
\end{equation}
where we replace $ \boldsymbol{w}_i\left( \boldsymbol{R}\boldsymbol{p}_i+\boldsymbol{t}-\boldsymbol{q}_i \right)  $ by $\boldsymbol{s}_i$ to simplify the formula. Due to the space limit, we directly present the iterative steps as
\begin{align}
    \boldsymbol{z}^{\left( k+1 \right)}                                     & :=\mathrm{arg}\min_{\boldsymbol{z}} L_{\rho}\left( \boldsymbol{R}^{\left( k \right)},\boldsymbol{t}^{\left( k \right)},\boldsymbol{z},\boldsymbol{\lambda }^{\left( k \right)} \right),\label{s1} \\
    \boldsymbol{R}^{\left( k+1 \right)},\boldsymbol{t}^{\left( k+1 \right)} & :=\mathrm{arg}\min_{\boldsymbol{R},\boldsymbol{t}} L_{\rho}\left( \boldsymbol{R},\boldsymbol{t},\boldsymbol{z}^{\left( k+1 \right)},\boldsymbol{\lambda }^{\left( k \right)} \right),\label{s2}   \\
    \boldsymbol{\lambda }_{i}^{\left( k+1 \right)}                          & :=\boldsymbol{\lambda }_{i}^{\left( k \right)}+\rho \left( \boldsymbol{z}_{i}^{\left( k \right)}-\boldsymbol{s}_{i}^{\left( k \right)} \right), \quad i = 1,\ldots, n,
\end{align}
where sub-problem~\eqref{s1} can be solved efficiently by the following shrinkage operator:
\begin{equation}
    \boldsymbol{z}_{i}^{\left( k+1 \right)}=\mathcal{S} _{1/\rho}\left( \boldsymbol{s}_{i}^{\left( k \right)}-\frac{1}{\rho}\boldsymbol{\lambda }^{\left( k \right)} \right) ,\qquad \mathcal{S} _{\lambda}\left( x \right) =\begin{cases}
        x-\lambda & \mathrm{if}\,x>\lambda \,; \\
        x+\lambda & \mathrm{if}\,x<\lambda \,; \\
        0         & \mathrm{otherwise}\,.      \\
    \end{cases},
\end{equation}
while sub-problem~\eqref{s2} can be solved by SVD.

\section{Experiments}\label{sec:experiment}
In this section, we compare the performance of the proposed method with three representative state-of-the-art approaches for multi-view registration, namely, JRMPC~\cite{evangelidis2017joint},
TMM~\cite{ravikumar_group-wise_2018}, and EMPMR~\cite{zhu_registration_2020}. The implementation of all compared methods is publicly available. All experiments in the following are performed on a laptop with a 6-core 2.2GHz Intel CPU and 16GB RAM.

\subsection{Data Sets and Evaluation Measure}
We use six 3D data sets from the Stanford 3D Scanning Repository\footnote{http://graphics.stanford.edu/data/3Dscanrep} (Bunny, Buddha, Dragon, and Armadillo) and the AIM@SHAPE Repository\footnote{http://visionair.ge.imati.cnr.it/ontologies/shapes} (Bimba and Olivier hand) for test. To quantitatively evaluate the registration performance of different methods, we compute the error by
\begin{equation}\label{eR}
    e_{\boldsymbol{R}}=\frac{1}{M}\sum_{i=1}^M{\mathrm{arc}\cos \left( \frac{\mathrm{tr}\left( \boldsymbol{R}_i\left( \boldsymbol{R}_{i}^{G} \right) ^T \right) -1}{2} \right)},\quad e_{\boldsymbol{t}}=\frac{1}{M}\sum_{i=1}^M{\left\| \boldsymbol{t}_i-\boldsymbol{t}_{i}^{G} \right\| _2},
\end{equation}
where $\left\{ \boldsymbol{R}_{i}^{G},\boldsymbol{t}_{i}^{G} \right\} _{i=1}^{M}$ and $\left\{ \boldsymbol{R}_{i},\boldsymbol{t}_{i} \right\} _{i=1}^{M}$ denote the ground truth and the estimated rigid transformation, respectively.

\subsection{Comparison Results}
We first downsample each data set to 4,000 points, and crop them along the $xy$-plane to generate the missing overlapping, then we rotate them around the $x$, $y$  and $z$-axes with the rotation angle uniformly distributed between $[-20^{\circ}, 20^{\circ}]$. Moreover, we add Gaussian noise to each data point with the signal-to-noise ratio (SNR) equal to 70dB and $30\%$ outliers following the uniform distribution.

\begin{figure}
    \centering
    \includegraphics[width=0.8\textwidth]{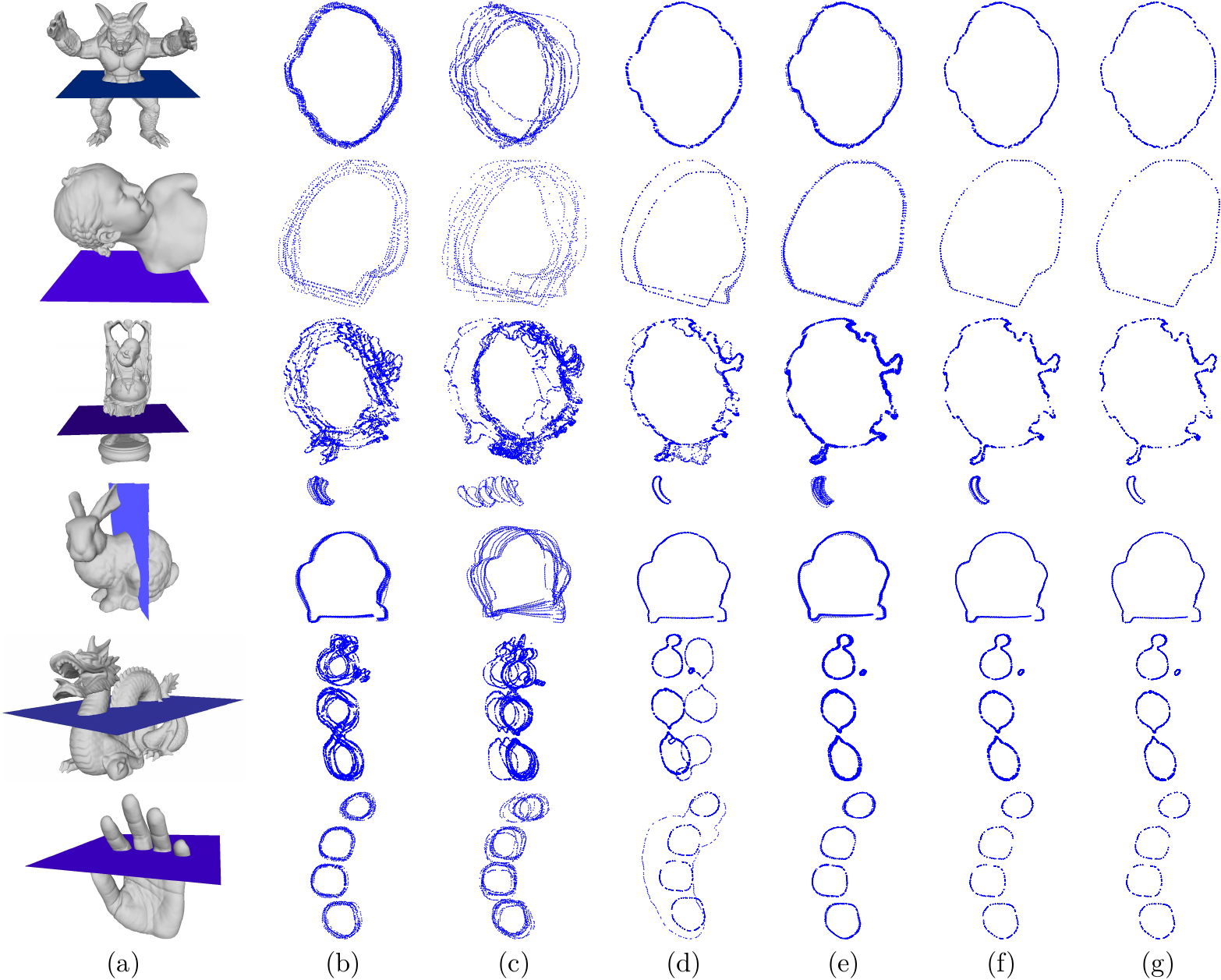}
    \vskip -0.3cm
    \caption{Sample registration results presented in the form of cross section. (a) 3D models. (b) Initial poses. (c) TMM.
        (d) JRMPC. (e) EMRPC . (f) Ours-ADMM. (g) Ours-LPA.}
    \label{fig:cross}
\end{figure}

The comparison results are reported in Table~\ref{tab:Comparison}. We color the best in \textcolor{red}{red} and the second-best in \textcolor{blue}{blue} for each metric. As observed, compared with initial errors, TMM has larger rotation deviations for all data sets, JRMPC and EMPMR also suffer from rotation estimation, such as the Bunny and the Armadillo data sets. In contrast, our proposed methods including Ours-ADMM and Ours-LPA achieve fewer rotation errors than initial cases for all data sets. Moreover, the proposed Ours-LPA attains the highest accuracy for rotation estimation than all competitors. EMPMR has the overall lowest translation error, but its rotation errors are relatively large. Our proposed method Ours-LPA attains the second highest accuracy for translation estimation, and the deviations are small enough. Thereby, the proposed method Ours-LPA achieves the overall best performance. We present several test examples in Fig.~\ref{fig:cross}.

\begin{table}[t]
    \centering
    \caption{Statistics of registration errors of all compared methods, where the \textcolor{red}{red} and \textcolor{blue}{blue} fonts indicate the best and the second-best performance for each metric. The proposed method Ours-LPA attains the overall best performance.}
    \setlength{\tabcolsep}{1.1mm}{
        \begin{tabular}{|c|c|c|c|c|c|c|c|}
            \hline
            \textbf{Methods}                                    &                      & Armadillo                  & Bimba                      & Buddha                     & Bunny                      & Dragon                     & Hand                       \\
            \hline
            \multirow{2}*{Initial}                              & $e_{\boldsymbol{R}}$ & 0.037378                   & 0.04016                    & 0.034194                   & 0.036776                   & 0.033269                   & 0.038187                   \\
                                                                & $e_{\boldsymbol{t}}$ & 0.000000                   & 0.000000                   & 0.000000                   & 0.000000                   & 0.000000                   & 0.000000                   \\
            \hline
            \multirow{2}*{TMM~\cite{ravikumar_group-wise_2018}} & $e_{\boldsymbol{R}}$ & 0.165546                   & 0.182914                   & 0.150696                   & 0.204346                   & 0.152766                   & 0.085944                   \\
                                                                & $e_{\boldsymbol{t}}$ & 0.001657                   & 0.003268                   & 0.009475                   & 0.008532                   & 0.011171                   & 0.002163                   \\
            \hline
            \multirow{2}*{JRMPC~\cite{evangelidis2017joint}}    & $e_{\boldsymbol{R}}$ & \textcolor{blue}{0.004748} & 0.067163                   & 0.024169                   & \textcolor{blue}{0.007340} & \textcolor{blue}{0.002504} & 0.312689                   \\
                                                                & $e_{\boldsymbol{t}}$ & 0.000087                   & 0.001867                   & 0.000615                   & 0.000246                   & 0.000259                   & 0.001234                   \\
            \hline
            \multirow{2}*{EMPMR~\cite{zhu_registration_2020}}   & $e_{\boldsymbol{R}}$ & 0.042555                   & 0.008452                   & 0.021825                   & 0.029602                   & 0.032349                   & 0.010282                   \\
                                                                & $e_{\boldsymbol{t}}$ & \textcolor{red}{0.000008}  & \textcolor{red}{0.000005}  & \textcolor{red}{0.000019}  & \textcolor{red}{0.000021}  & \textcolor{red}{0.000028}  & \textcolor{red}{0.000003}  \\
            \hline
            \multirow{2}*{Ours-ADMM}                            & $e_{\boldsymbol{R}}$ & 0.008538                   & \textcolor{blue}{0.001002} & \textcolor{blue}{0.004597} & 0.008356                   & 0.004588                   & \textcolor{blue}{0.002335} \\
                                                                & $e_{\boldsymbol{t}}$ & 0.000120                   & 0.000075                   & 0.000301                   & 0.000713                   & 0.000326                   & 0.000080                   \\
            \hline
            \multirow{2}*{Ours-LPA}                             & $e_{\boldsymbol{R}}$ & \textcolor{red}{0.002954}  & \textcolor{red}{0.000353 } & \textcolor{red}{0.001519 } & \textcolor{red}{0.001953}  & \textcolor{red}{0.001181}  & \textcolor{red}{0.000896}  \\
                                                                & $e_{\boldsymbol{t}}$ & \textcolor{blue}{0.000051} & \textcolor{blue}{0.000019} & \textcolor{blue}{0.000114} & \textcolor{blue}{0.000159} & \textcolor{blue}{0.000101} & \textcolor{blue}{0.000023} \\
            \hline
        \end{tabular}%
    }
    \label{tab:Comparison}%
\end{table}%

We further report the time consumption of different methods in Table~\ref{tab:time}. As observed, due to the closed-form solution of EMPMR based on the $L_2$ norm, it consumes relatively less time. Our proposed method Ours-ADMM has the second fastest speed, and its time consumption is quite close to EMPMR. Moreover, from Table~\ref{tab:Comparison}, we find that Ours-ADMM has fewer rotation errors than EMPMR, meanwhile with acceptable translation deviations. Ours-LPA consumes relatively more time than Ours-ADMM, however, it is still faster than TMM and JRMPC. Therefore, to register point sets with high efficiency and comparable accuracy, we suggest Ours-ADMM for use.

\subsection{Robustness against Outliers}
\begin{figure}
    \centering
    \subfigure[Bunny model]{
        \includegraphics[width=0.31\textwidth]{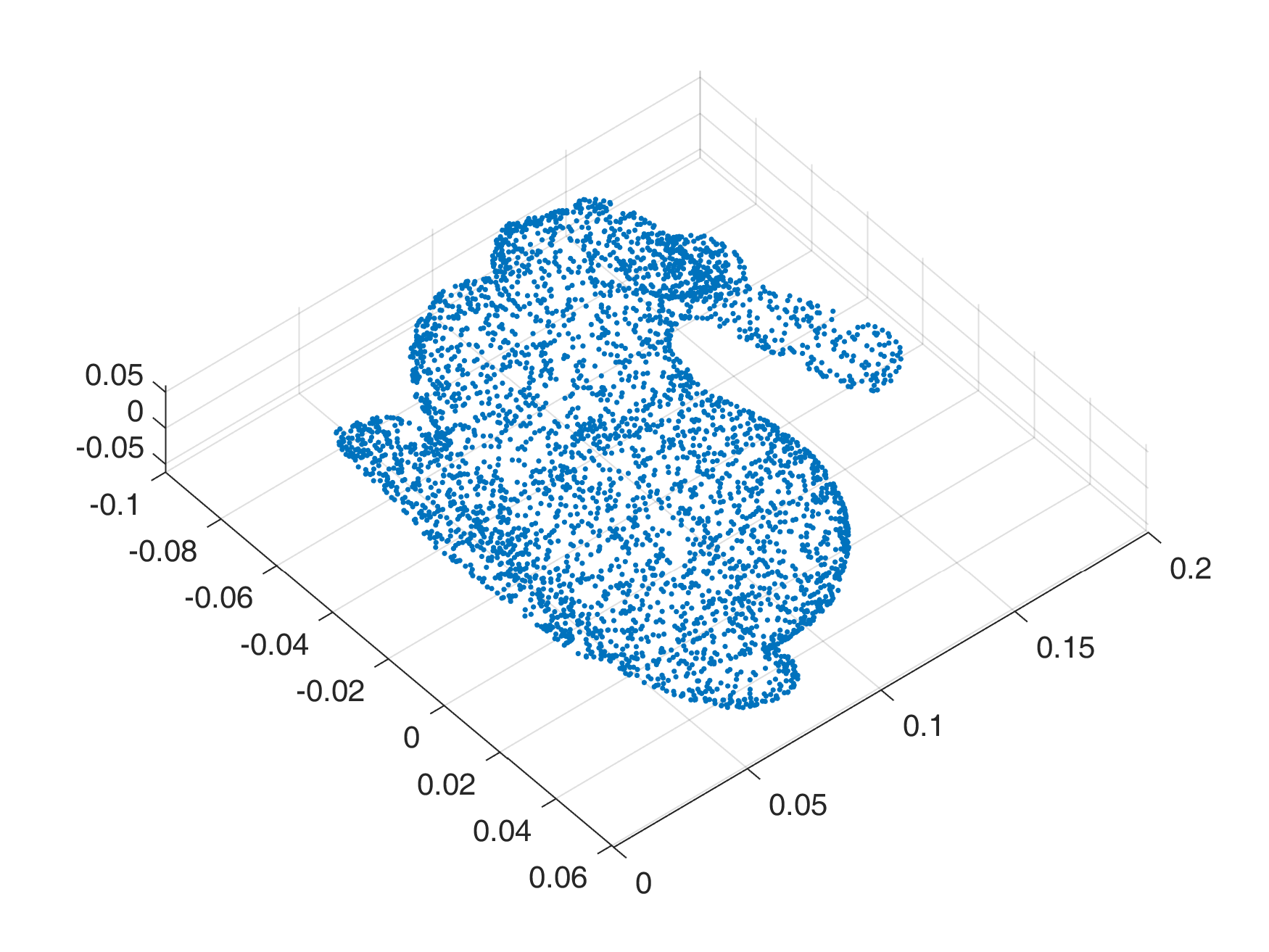}
    }
    \subfigure[SNR=50dB]{
        \includegraphics[width=0.31\textwidth]{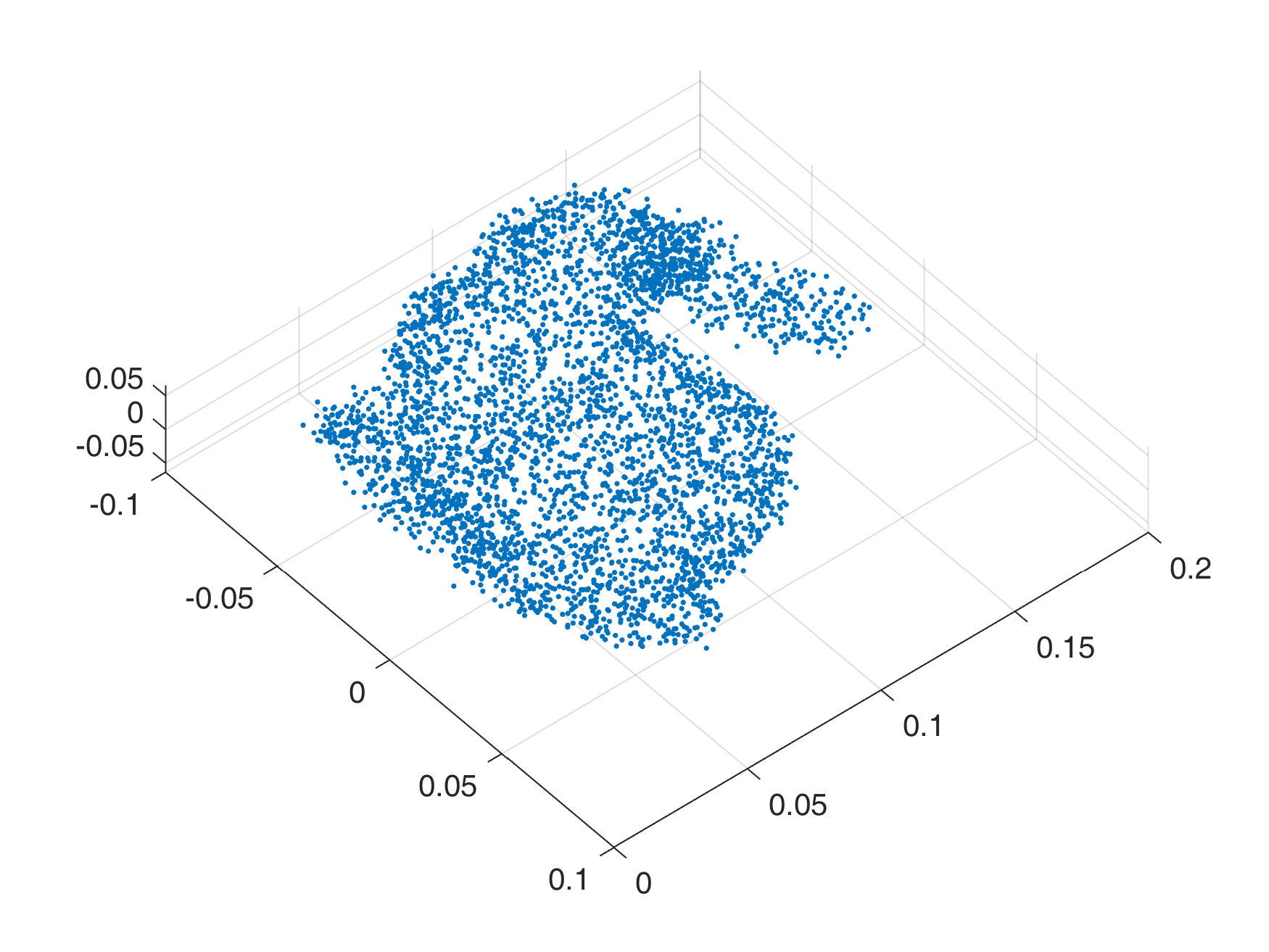}
    }
    \subfigure[30\% outliers]{
        \includegraphics[width=0.31\textwidth]{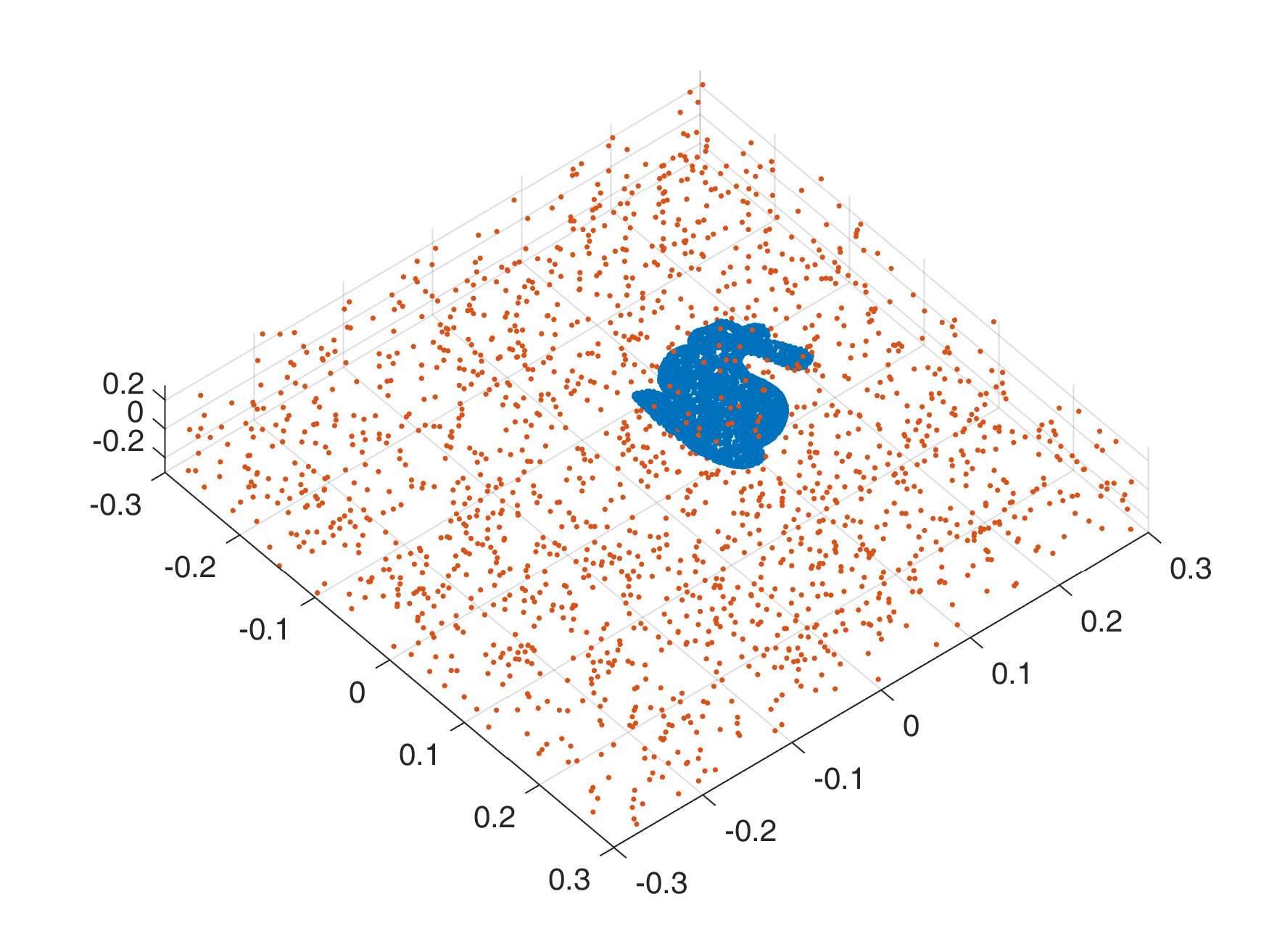}
    }
    \vskip -0.3cm
    \caption{Examples of the Bunny model contaminated by noise and outliers.}
    \label{fig:model}
    \vskip -0.3cm
\end{figure}

Subsequently, we evaluate the robustness of the proposed method against outliers. We adopt the bunny data set for this purpose. We add $1\%-80\%$ outliers to the point cloud. Besides, we contaminate the point cloud with 70dB Gaussian noise. Several examples are illustrated in Fig.~\ref{fig:model}. The test results are reported in Table~\ref{tab:outliers}. As can be seen, with outlier increasing, the proposed method Ours-LPA exhibits higher robustness than compared ones. EMPMR has fewer deviations at low outlier contamination, but it suffers from severe outliers (more than 30\%). JRMPC also shows certain robustness against outliers, whereas its performance is unstable. In contrast, Ours-ADMM attains the quite stable performance for all outlier tests, and it has the a very similar registration accuracy to the first two winners. TMM has the largest deviations than the others, indicating its weakness in handling outliers.
\begin{table}[t]
    \centering
    \caption{Time consumption of all compared methods.}
    \begin{tabular}{ccccccc}
        \toprule
        Methods                              & Armadillo   & Bimba      & Buddha      & Bunny      & Dragon      & Hand       \\
        \midrule
        TMM~\cite{ravikumar_group-wise_2018} & 416.048315  & 306.375451 & 428.698706  & 215.150639 & 456.21449   & 328.926923 \\

        JRMPC~\cite{evangelidis2017joint}    & 260.635473  & 234.189799 & 315.937198  & 69.655174  & 73.304373   & 163.161366 \\

        EMPMR~\cite{zhu_registration_2020}   & 14.298502   & 7.159238   & 10.059491   & 5.108593   & 6.233101    & 4.700202   \\

        Ours-ADMM                            & {21.439405} & {8.915458} & {51.801415} & {8.883673} & {26.442176} & {8.690539} \\

        Ours-LPA                             & 108.499447  & 36.490062  & 161.175015  & 20.262704  & 96.782577   & 36.421847  \\
        \bottomrule
    \end{tabular}%
    \label{tab:time}%
\end{table}%

\begin{table}[htbp]
    \centering
    \caption{Statistics of registration errors under the contamination of different outliers (\%). The proposed method Ours-LPA has the overall best performance.}
    \setlength{\tabcolsep}{2mm}{
        \begin{tabular}{ccccccc}
            \toprule
                                                 & 1\%                        & 10\%                       & 20\%                       & 30\%                       & 50\%                       & 80\%                       \\
            \midrule
            Initial                              & 0.032794                   & 0.034169                   & 0.035956                   & 0.036776                   & 0.028585                   & 0.027937                   \\

            TMM~\cite{ravikumar_group-wise_2018} & 0.010005                   & 0.157573                   & 0.162055                   & 0.204346                   & 0.269017                   & 0.293706                   \\

            JRMPC~\cite{evangelidis2017joint}    & \textcolor{blue}{0.004102} & 0.299748                   & 0.009423                   & \textcolor{blue}{0.007340} & \textcolor{blue}{0.005685} & 0.077505                   \\

            EMPMR~\cite{zhu_registration_2020}   & \textcolor{red}{0.000342}  & \textcolor{blue}{0.000378} & \textcolor{red}{0.000500}  & 0.029602                   & 0.052858                   & 0.094169                   \\

            Ours-ADMM                            & 0.000439                   & 0.000572                   & 0.008176                   & 0.008356                   & 0.007888                   & \textcolor{blue}{0.034256} \\

            Ours-LPA                             & 0.000505                   & \textcolor{red}{0.000369}  & \textcolor{blue}{0.000676} & \textcolor{red}{0.001953}  & \textcolor{red}{0.002762}  & \textcolor{red}{0.021861}  \\
            \bottomrule
        \end{tabular}%
    }
    \label{tab:outliers}%
\end{table}%

\subsection{Robustness against Noise}
We also test the performance of the proposed methods in term of noise. To this end, we first contaminate the point cloud with 30\% outliers, and then increase the noise intensity by decreasing the signal-to-noise ratio. The test results are reported in Table~\ref{tab:noises}. As observed, the proposed method Ours-LPA achieves the overall best performance, and its registration errors are even 1,000 times fewer than TMM. JRMPC and Ours-ADMM attain similar results, thereby they have the second-best performance. EMPMR shows shortcomings for noise, especially for low SNR noise.

\begin{table}[htb]
    \vskip -0.4cm
    \centering
    \caption{Statistics of registration errors of different methods under the contamination of different noise (dB).}
    \setlength{\tabcolsep}{3mm}{
        \begin{tabular}{ccccccc}
            \toprule
                                                 & 90                         & 80                         & 70                         & 60                         & 50                         \\
            \midrule
            Initial                              & 0.035592                   & 0.028818                   & 0.036776                   & 0.029031                   & 0.035445                   \\

            TMM~\cite{ravikumar_group-wise_2018} & 0.159936                   & 0.163456                   & 0.204346                   & 0.179884                   & 0.162614                   \\

            JRMPC~\cite{evangelidis2017joint}    & 0.093772                   & \textcolor{red}{0.000439 } & \textcolor{blue}{0.007340} & 0.065524                   & \textcolor{blue}{0.020319} \\

            EMPMR~\cite{zhu_registration_2020}   & 0.039599                   & 0.056480                   & 0.029602                   & 0.033342                   & 0.056580                   \\

            Ours-ADMM                            & \textcolor{blue}{0.008533} & 0.008808                   & 0.008356                   & \textcolor{blue}{0.009525} & \textcolor{red}{0.019820}  \\

            Ours-LPA                             & \textcolor{red}{0.000245}  & \textcolor{blue}{0.001216} & \textcolor{red}{0.001953}  & \textcolor{red}{0.007375}  & 0.032890                   \\
            \bottomrule
        \end{tabular}%
    }
    \label{tab:noises}%
    \vskip -0.6cm
\end{table}%

\section{Conclusion}
We have presented a novel and robust multi-view registration method for point clouds based on the Laplacian mixture models (LMM). We adopt Laplacian distribution to represent each data point, and then cast the multi-view registration task as a density estimation problem, which can be efficiently solved through the expectation-maximization framework. Due to the heavy tail and the sparsity-induced $L_1$ norm, LMM is more robust against GMM. To solve the $L_1$ problem, we deduce our objective function into two optimization paradigms, namely, linear programming and ADMM. We test the proposed methods on  challenging data sets with contamination of noise and outliers, and compare it with three representative state-of-the-art approaches, and results demonstrate the salient advantages of the proposed method: more robust against noise and outliers, as well as higher accuracy.

\subsubsection*{Acknowledgements}

This work was supported by the National Key Research \& Development Program of China (2020YFA0713701), the National Natural Science Foundation of China (12171023 and 62172415), and the Open Research Fund Program of State Key Laboratory of Hydroscience and Engineering, Tsinghua University (sklhse-2020-D-07).

%
%
%
\bibliographystyle{splncs04}
\bibliography{references}

\end{document}